%% file: acl2020.tex
\documentclass[11pt,a4paper]{article}
\usepackage[dvipsnames]{xcolor}
\usepackage{acl}
\usepackage{times}
\usepackage{latexsym}
\usepackage{booktabs}
\usepackage{multirow}
\usepackage{blindtext}
\usepackage{subcaption}
\usepackage{amsmath,amssymb}
\usepackage{graphicx}
\usepackage{url}
\usepackage{float}
\usepackage{bm}
\usepackage{scalerel}
\usepackage[size=tiny]{todonotes}
\usepackage[normalem]{ulem}
\usepackage[linesnumbered,ruled]{algorithm2e}
\useunder{\uline}{\ul}{}
\usepackage{enumitem}

\usepackage{microtype}

\title{Layer or Representation Space:\\ What makes BERT-based {Evaluation} Metrics Robust?}

\author{%
Doan Nam Long Vu$^1$, Nafise Sadat Moosavi$^2$, Steffen Eger$^3$\\ \\
$^1$ Department of Computer Science, Technical University of Darmstadt, Germany \\
$^2$ Department of Computer Science, The University of Sheffield, UK \\
$^3$ NLLG, Faculty of Technology, Bielefeld University, Germany \\
        \url{doannamlong.vu@stud.tu-darmstadt.de}
}

\date{}

\begin{document}
\maketitle
\begin{abstract}
The evaluation of recent embedding-based evaluation metrics for text generation is primarily based on measuring their correlation with human evaluations on standard benchmarks. However, these benchmarks are mostly from similar domains to those used for pretraining word embeddings. This raises concerns about the (lack of) generalization of embedding-based metrics to new and noisy domains that contain a different vocabulary than the pretraining data. In this paper, we examine the robustness of BERTScore, one of the most popular embedding-based metrics for text generation. We show that (a) an embedding-based metric that has the highest correlation with human evaluations on a standard benchmark can have the lowest correlation if the amount of input noise or unknown tokens increases, (b) taking embeddings from the first layer of pretrained models improves the robustness of all metrics, and (c) the highest robustness is achieved when using character-level embeddings, instead of token-based embeddings, from the first layer of the pretrained model.\footnote{The code of our experiments is available at \url{https://github.com/long21wt/robust-bert-based-metrics}}    
\end{abstract}

\input{introduction.tex} \label{introduction}
\input{approach.tex} \label{approach}
\input{experiments} \label{experiments}
\input{conclusion.tex}

\bibliography{acl2020}
\bibliographystyle{acl_natbib}
\input{appendix.tex}
\end{document}

%% file: introduction.tex
\section{Introduction}
Evaluating the quality of generated outputs by Natural Language Generation (NLG) models is a challenging and open problem. 
Human judgments can directly assess the quality of generated texts \citep{popovic-2020-informative,escribe-2019-human}. However, human evaluation, either with experts or crowdsourcing, is expensive and time-consuming. Therefore, automatic evaluation metrics, which are fast and cheap, are commonly used 
alternatives for the rapid development of text generation systems \citep{van-der-lee-etal-2019-best}. Traditional metrics such as BLEU \citep{papineni-etal-2002-bleu}, METEOR \citep{banerjee-lavie-2005-meteor}, and ROUGE \citep{lin-2004-rouge} measure $n$-gram overlap between generated and reference texts. While these metrics are easy to use, they cannot correctly assess generated texts that contain novel words or a rephrasing of the reference text.

Recent metrics like BERTScore \citep{bert-score}, MoverScore \citep{zhao-etal-2019-moverscore}, COMET \citep{rei-etal-2020-comet}, BARTScore \citep{yuan2021bartscore}, and BLEURT \citep{sellam-etal-2020-bleurt} adapt pretrained contextualized word embeddings to tackle this issue. 
These novel metrics have shown higher correlations with human judgments on various tasks and datasets \citep{ma-etal-2019-results,mathur-etal-2020-results}. However, the correlations are measured on standard benchmarks containing text domains similar to those used for pretraining the embeddings themselves. As a result, it is unclear how reliable these metrics are on domains and datasets containing words outside the vocabulary of the pretraining data.

The goal of this paper is to investigate the \emph{robustness} of embedding-based evaluation metrics on new and noisy domains that contain a higher ratio of unknown tokens compared to standard text domains.\footnote{We connect to recent research that investigates the behavior of metrics in adversarial situations \citep{sai-etal-2021-perturbation,kaster-etal-2021-global,Leiter2022TowardsEE,zeidler-etal-2022-dynamic}.} 
We examine the robustness of BERTScore, one of the most popular recent metrics for text generation.\footnote{E.g., as of September 2022, BERTScore is cited $\sim$1200 times while it is $\sim$200 and 400 for MoverScore and BLEURT, respectively.} 
In order to perform a systematic evaluation on the robustness of BERTScore with regard to the ratio of unknown tokens, we use character-based adversarial attacks \citep{eger2020hero} that introduce a controlled ratio of new unknown tokens to the input texts. Our contributions are: 

\begin{itemize}[topsep=5pt,itemsep=0pt,leftmargin=*]
   \item We investigate whether the use of character-based embeddings instead of token-based embeddings improves the robustness of embedding-based generation metrics. Our results show that the evaluations based on character-level embeddings are more robust.
   \item We examine the impact of the hidden layer used for computing the embeddings in BERTScore.  We show that the choice of hidden layer affects the robustness of the evaluation metric.
   \item We show that by using \textbf{character-level embeddings from the first layer}, we achieve the highest robustness, i.e., similar correlation with human evaluations for different ratios of unknown tokens.
\end{itemize}

%% file: approach.tex
\section{BERTScore}

BERTScore \citep{bert-score} computes the pairwise cosine similarity between the reference and hypothesis using contextual embeddings. It forward-passes sentences through a pretrained model, i.e., BERT \citep{devlin-etal-2019-bert}, and extracts the embedding information from a specific hidden layer. To select the best hidden layer, BERTScore uses average Pearson correlation with human scores on WMT16 \citep{bojar-etal-2016-results} over five language pairs. For instance, the best layer is the ninth layer for $\mathsf{BERT}_{\mathsf{base-uncased}}$. 

\paragraph{BERTScore with character-level embeddings.} Existing embedding-based metrics, including BERTScore, use token-based embeddings that are taken from pretrained models like BERT \citep{devlin-etal-2019-bert}. In this paper, we investigate the impact of using character-level embeddings instead of token-level embeddings in BERTScore \citep{bert-score}. We use ByT5 \citep{xue2021byt5}, which encodes the input at the byte level.
It tokenizes a word into a set of single characters or converts it directly to UTF-8 characters before forwarding the input sequence into the model. \citet{xue2021byt5} show that ByT5 is more robust to noise compared to word-level embeddings. For computing BERTScore using character-level embeddings, we use ByT5 instead of BERT in BERTScore computations. We adapt three variants of ByT5 (small, base, large) in BERTScore. Table~\ref{tab:byt5} presents the best layer of ByT5 models for computing BERTScore.

\begin{table}[!htb]
    \centering
    \footnotesize
    \begin{tabular}{@{}l|r|r@{}}
    \toprule
    Model      & Best Layer & Score \\ \midrule
    ByT5-{small} & 1           & 0.510 \\
    ByT5-{base}  & 17          & 0.581 \\
    ByT5-{large} & 30          & 0.615 \\ \bottomrule
    \end{tabular}
    \caption{Best layers with different ByT5 variants and their average Pearson correlation score on WMT16.}
    \label{tab:byt5}
\end{table}

%% file: experiments.tex
\section{Experimental settings}

\subsection{Evaluation on a standard benchmark}  
We report the results on the WMT19 dataset \citep{ma-etal-2019-results} that contains seven to-English language pairs. Each language pair has 2800 sentences, each corresponding to one reference, plus the systems' output sentences.
Totally, the human evaluation in WMT19 has 281k segment sample scores for each of the output translation in to-English language pairs. Table \ref{tab:wmt19-stat} shows the language pairs considered, as well as the number of segments per language pair.

\begin{table}[!tb]
  \centering
  {\small
  \begin{tabular}{@{}l|c@{}}
  \toprule
                                   Language Pairs             & No. Segment Sample (DARR) \\ \midrule
                                   de-en (German-English)     & 85365                     \\
                                   fi-en (Finnish-English)    & 38307                     \\
                                   gu-en (Gujarati-English)   & 31139                     \\
                                   kk-en (Kazakh-English)     & 27094                     \\
                                   lt-en (Lithuanian-English) & 21862                     \\
                                   ru-en (Russian-English)    & 46172                     \\
                                   zh-en (Chinese-English)    & 31070                     \\ \bottomrule
  \end{tabular}}
  \caption{To-English language pairs of WMT19. DARR denotes \emph{Direct Assessment Relative Ranks}, in which all available sentence pairs of DA (Direct Assessment) scores are taken into account.}
  \label{tab:wmt19-stat}
  \end{table}

\subsection{Evaluating Robustness}
\paragraph{Evaluation on different ratios of unknown tokens.} To evaluate the robustness of 
{evaluation} metrics on new domains, we use character-level attacks to introduce a controlled ratio of unknown tokens in the corresponding reference texts of the evaluation sets.\footnote{We need human annotations for evaluating the correlation of evaluation metrics with human judgments, and such annotations are available for standard domains like WMT datasets. As a result, we introduce unknown tokens by using character-level attacks to artificially introduce more unknown tokens.} 
We examine five different attacks from \citet{eger2020hero}: (a) \textbf{intruders:} inserting a character---e.g., \textsf{`.', `/', `:'}---in between characters of a word,
 (b) \textbf{disemvoweling}: removing vowels---e.g., `a', `e', `i'---from the word, (c) \textbf{keyboard typos}: randomly replacing letters of a word with characters that are 
nearby the original characters on an English keyboard, (d) \textbf{phonetic}: changing a word's spelling in such a way that its pronunciation stays the same,  
and (e) \textbf{visual}: replacing characters with a symbol that is its visually nearest neighbor \citep{eger2019text}.
We can control the ratio of tokens that are affected by the adversarial attack by the \emph{perturbation level} ($p$), e.g., $p=0$ denotes no attack and $p=0.3$ indicates that each letter in the sentence is attacked by the probability of 0.3. Table~\ref{tab:attack} shows an example of each of these attacks at $p=0.3$.

\begin{table}[!tb]
    \centering
    {\footnotesize
    \begin{tabular}{@{}l|l@{}}
    \toprule
    Setting       & Sentence                                                           \\ \midrule
    no-attack     & Now they have come to an agreement.                                \\
    intrude       & Now they have c/o/me t+o a\textgreater{}n agreement.               \\
    disemvowel    & Nw thy have come to an grmnt.                                      \\
    keyboard-typo & No3 they have come to xn agrrement.                                \\
    phonetic      & Nau they have cohm to an agrimand.                                 \\
    visual        & \scalerel*{\includegraphics{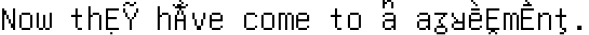}}{15pt} \\
    \bottomrule
    \end{tabular}}
    \caption{Examples for the character-level attacks \citep{eger2020hero,keller-etal-2021-bert} at perturbation level $p=0.3$, i.e., the probability that each letter in a sentence is attacked is 0.3.}
    \label{tab:attack}
\end{table}

\begin{figure}[!tb]
  \centering
  \includegraphics[width=\linewidth]{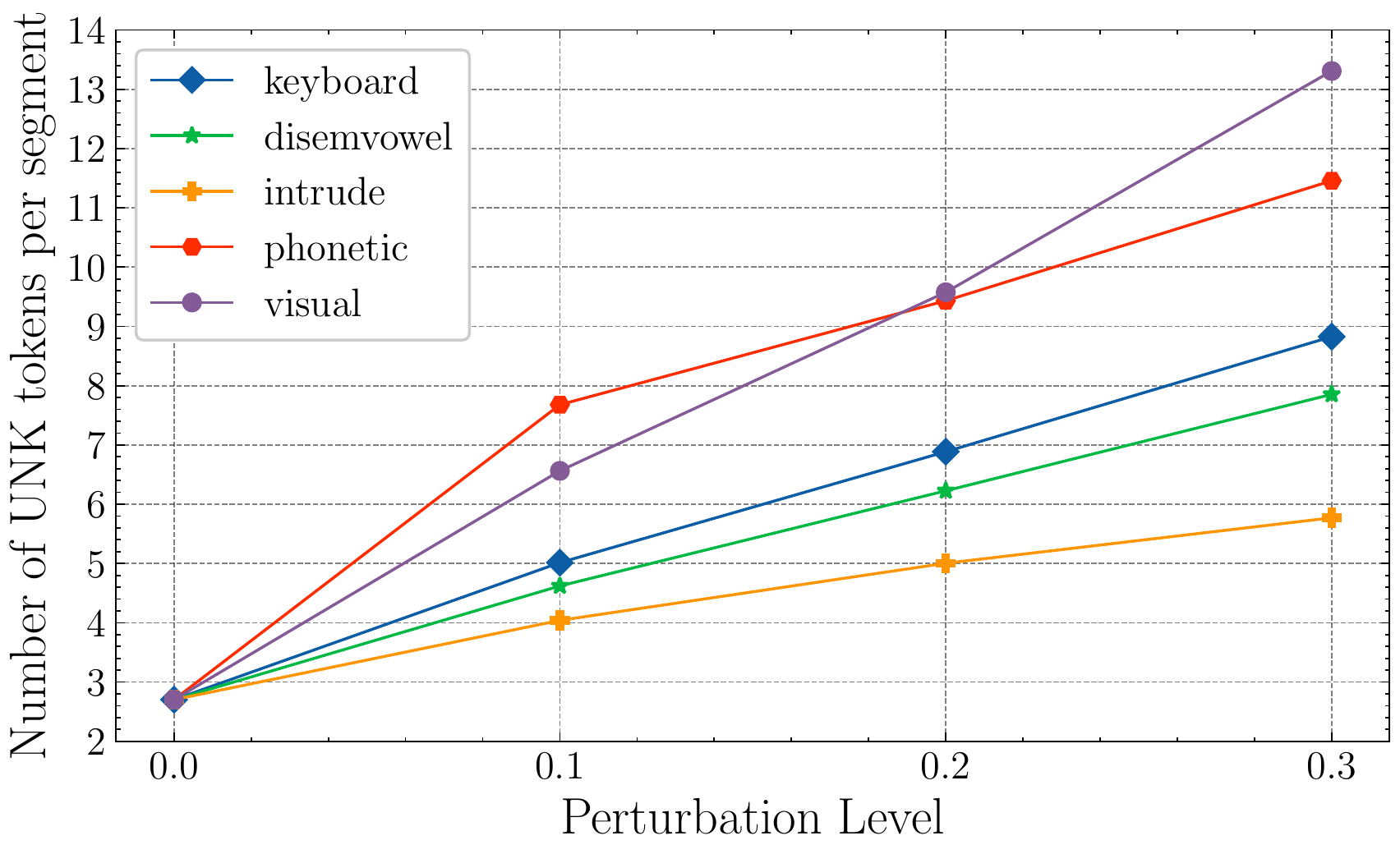}
  \caption{The number of average unknown tokens per segment across seven to-English language pairs in WMT19 given different attacks and perturbation levels.}
  \label{fig:wmt19-unk}
\end{figure}

Figure~\ref{fig:wmt19-unk} shows the average number of unknown tokens, as determined based on BERT's tokenizer, per segment across seven to-English language pairs given different attacks and perturbation levels. We count a token as an unknown token if (1) BERT represents it as [UNK], or (2) BERT splits it into subwords, e.g., \emph{`pre-trained'} to \emph{`pre',`\#\#train',`\#\#ed'}.\footnote{Please refer to the detailed algorithm in Appendix~\ref{sect:label_count}.}
As we see from the figure, the number of unknown tokens increases as we apply these character-level attacks with higher perturbation levels. In our experiments in Section~\ref{sect:exp}, we report the results using visual attacks. The results using other attacks are also reported in Appendix \ref{sect:wmt19}, and they follow the same patterns as those using the visual attack. 

\paragraph{Evaluation on low-resource language pairs.} Apart from the experiments on WMT19, we also perform the evaluations on the (Xhosa, Zulu) and (Bengali, Hindi) language pairs from WMT21 \citep{freitag2021results}. BERTScore uses multilingual BERT for evaluating non-English languages. Multilingual models contain a higher ratio of unknown tokens for low-resource languages, and therefore, evaluating the correlation of embedding-based metrics with human judgments on low-resource languages is also an indicator of their robustness. Table \ref{tab:unk-low-resources} shows the number of unknowns tokens per segment to multilingual BERT in four different low-resources language pairs in WMT21 dataset. We refer to the number of segments of low-resources dataset in Table \ref{tab:flores} in Appendix \ref{sect:flores}.

\begin{table}[!tb]
\centering
\begin{tabular}{@{}l|r@{}}
\toprule
Language pair         & No. unknown tokens \\ \midrule
bn-hi (Bengali-Hindi) & 19.235                \\
hi-bn (Hindi-Bengali) & 23.478               \\
xh-zu (Xhosa-Zulu)    & 28.930               \\
zu-xh (Zulu-Xhosa)    & 28.743               \\ \bottomrule
\end{tabular}
\caption{The number of average unknown tokens per segment for each language pair in our low-resource datasets.}
\label{tab:unk-low-resources}
\end{table}

\section{Experiments}
\label{sect:exp}
\subsection{Impact of Character-level Embeddings}
\label{sect:res}
\begin{table*}[!htb]
    \centering
    \footnotesize
    \begin{tabular}{@{}lcccccccc@{}}
    \toprule
                                               & de-en          & fi-en          & gu-en          & kk-en          & lt-en          & ru-en          & zh-en          & Average\\ \midrule
    BERT-base                          & 0.180          & 0.339          & 0.288          & 0.438          & 0.364          & 0.209          & 0.410          & 0.318\\
    BERT-large                         & 0.194          & \textbf{0.346} & 0.292          & \textbf{0.442} & \textbf{0.375} & 0.208          & \textbf{0.418} & \textbf{0.325}\\
    ByT5-small                                 & 0.172          & 0.286          & 0.278          & 0.422          & 0.307          & 0.194          & 0.373          & 0.290\\
    ByT5-base                                  & \textbf{0.197} & 0.326          & 0.297          & 0.419          & 0.358          & \textbf{0.215} & \textbf{0.418} & 0.319\\ 
    ByT5-large                                 & 0.193          & 0.333          & \textbf{0.304} & 0.427          & 0.354          & 0.208          & 0.415          & 0.319\\\bottomrule
    \end{tabular}
    \caption{Segment-level Kendall correlation results for to-English language pairs in WMT19 without any attack, i.e. $p=0$. The correlation of BERTScore with human are reported using different embeddings including bert-base-uncased, bert-large-uncased, ByT5-small, ByT5-base, and ByT5-large. }
    \label{wmt19-to-en-no-attack}
\end{table*}

\begin{figure}[!tb]
    \centering
    \includegraphics[width=\linewidth]{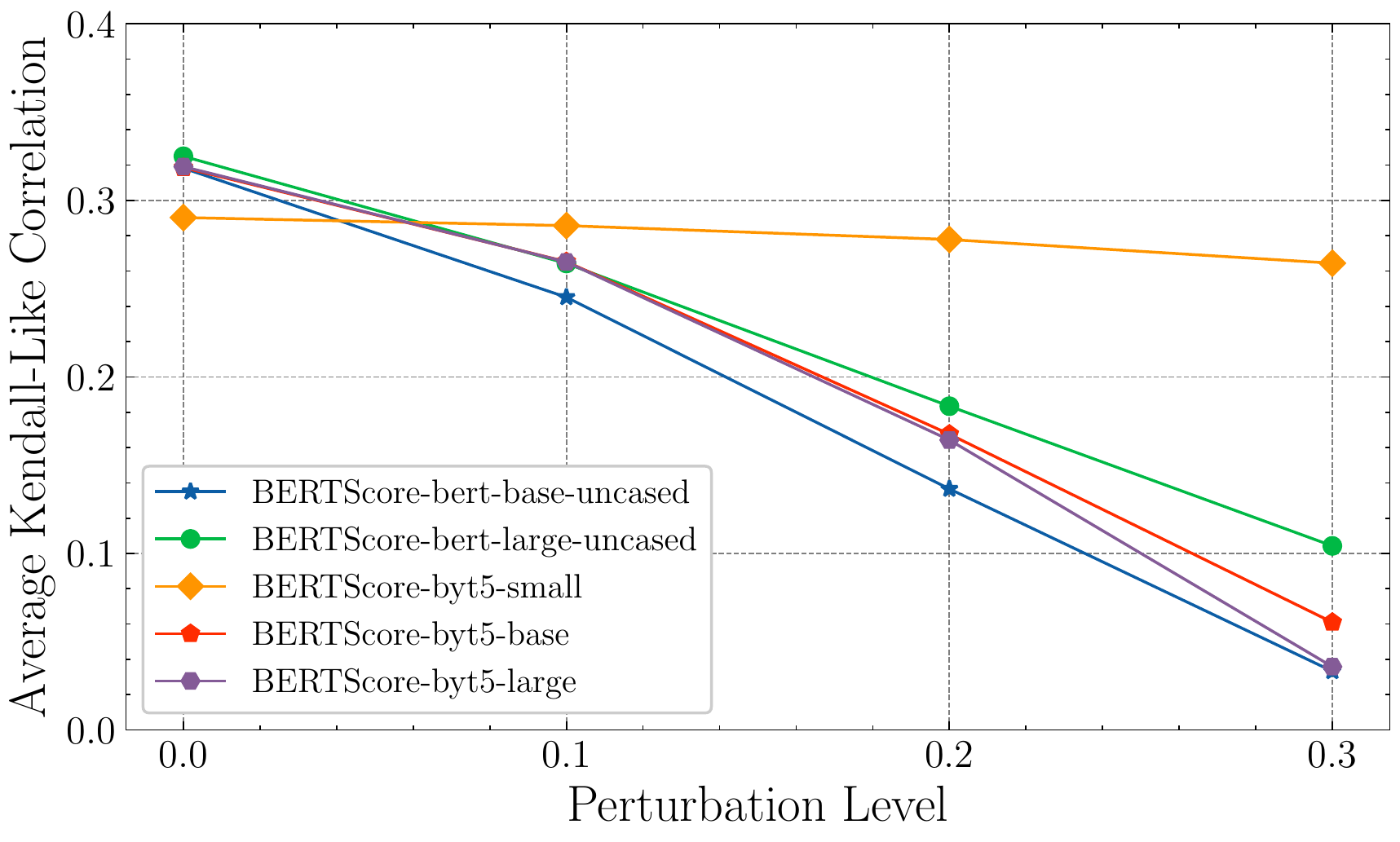}
  \caption{Average Kendall correlation of 7 to-English language pairs in WMT19 given different perturbation level from $p=0.0$ to $p=0.3$ using the visual attack.}
  \label{fig:wmt19-visual}
\end{figure}

Table~\ref{wmt19-to-en-no-attack} shows the results of BERTScore using different embeddings on WMT19's to-English language pairs {(using $p=0$)}. 
Figure~\ref{fig:wmt19-visual} shows the average correlation score over all seven to-English language pairs given different perturbation level from $p=0$ to $p=0.3$ using the visual attack.

We observe that computing BERTScore using the ByT5-small models results in a slightly lower average correlation with human scores over the seven to-English pairs at $p=0$ compared to BERTScore using BERT and larger ByT5 models.

However, the average correlation using ByT5-small remains around the same value given different ratio of unknown tokens,  indicating higher \emph{robustness} of the metrics using ByT5-small. 
On the other hand, while using BERT-large embeddings results in the highest average correlation with human scores in Table~\ref{wmt19-to-en-no-attack}, its correlation drops considerably in the presence of more unknown tokens in Figure~\ref{fig:wmt19-visual}.

\begin{figure*}[!tb]
  \centering
  \begin{subfigure}{0.5\linewidth}
    \centering
    \includegraphics[width=\linewidth]{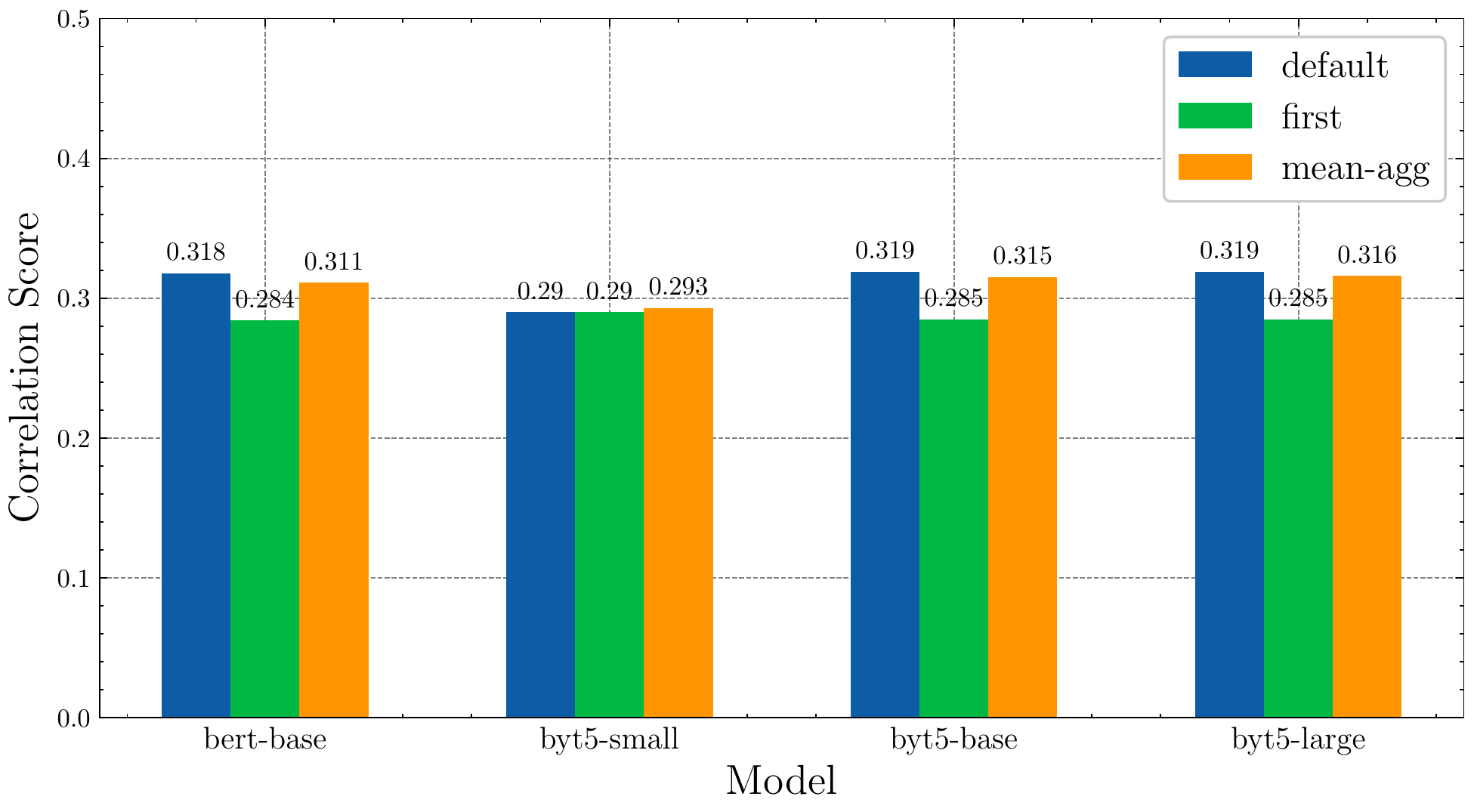}
    \caption{No attack}
  \end{subfigure}%
  \begin{subfigure}{0.5\linewidth}
    \centering
    \includegraphics[width=\linewidth]{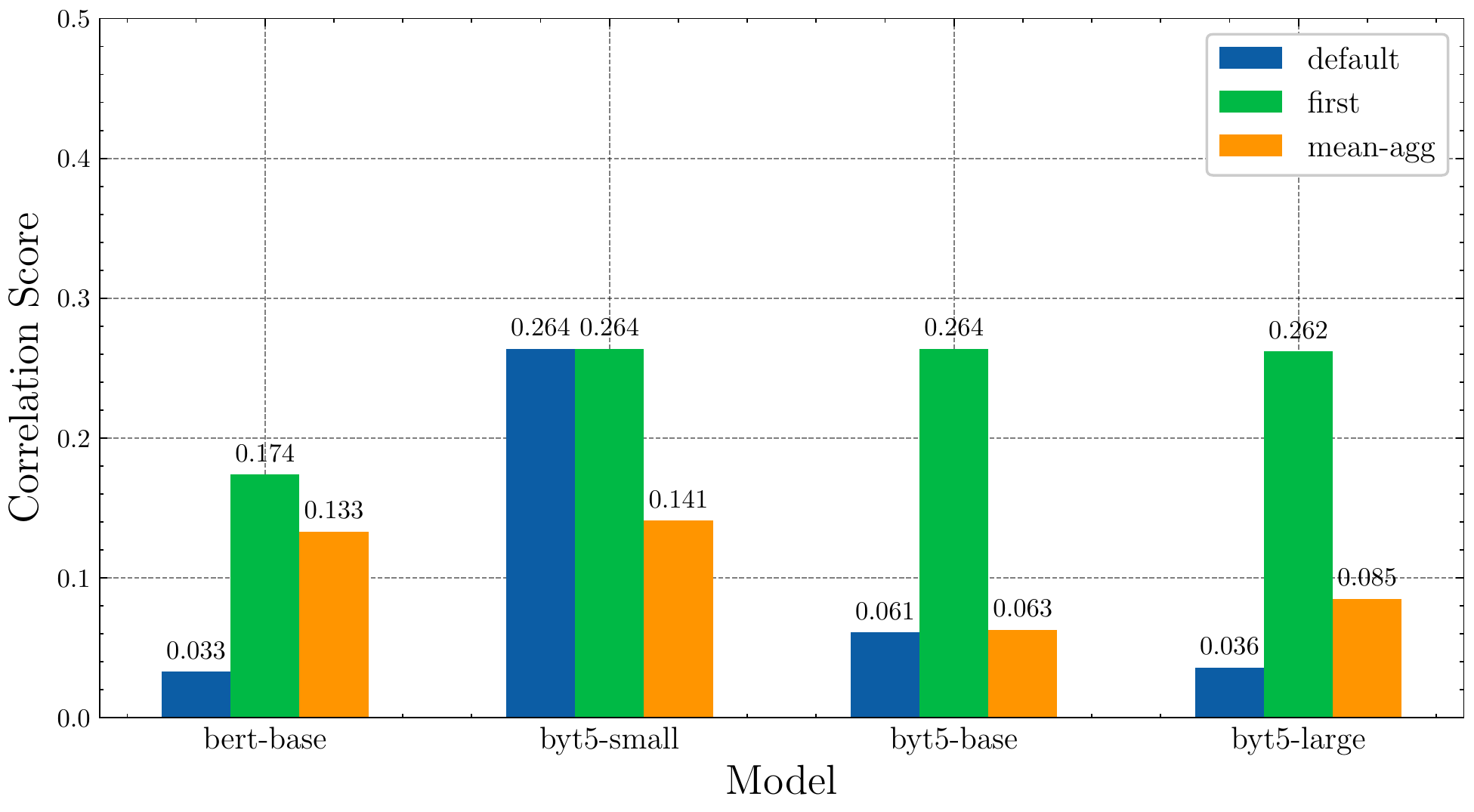}
    \caption{Visual attack at $p=0.3$}
  \end{subfigure}
  \caption{Average segment-level Kendall correlation results  for seven to-non-English language pairs in WMT19 to fist layer, default layer, and mean of aggregated embeddings setting in BERTScore.}
  \label{fig:why}
\end{figure*}

For Hindi-Bengali and Zulu-Xhosa, we compare the results against using the BERT-base-multilingual model in Table~\ref{tab:low_resource}. 
We observe that the BERTScore metric that uses ByT5-small achieves higher correlations with humans throughout. Given that low resources languages contain more out-of-vocabulary words for pretrained models, this observation confirms our previous results using character-level attacks on the WTM19 dataset.
\subsection{Impact of the Selected Hidden Layer}

\begin{table}[t]
    \centering
    \footnotesize
    \begin{tabular}{@{}lllll@{}}
    \toprule
    Model      & bn-hi & hi-bn  & xh-zu & zu-xh \\ \midrule
    BERT-multi & 0.073  & 0.364 & 0.266 & 0.488\\
    ByT5-{small} &  \textbf{0.096} & \textbf{0.411} & \textbf{0.311} & \textbf{0.523}  \\
    \bottomrule
    \end{tabular}
    \caption{Kendall correlation scores of BERTScore for WMT21 low-resource language pairs Hindi-Bengali and Zulu-Xhosa using BERT-base-multilingual and ByT5-small embeddings.}
    \label{tab:low_resource}
\end{table}

Our results in Section~\ref{sect:res} show the robustness of BERTScore when using the ByT5-small model for computing the embeddings. 
However, as Table~\ref{tab:byt5} shows, the selected hidden layer for getting embeddings varies when using different pretrained models. For instance, when using ByT5-small embeddings, the model uses the embeddings of the first layer while it uses the embeddings of the 30th layer for ByT5-large.
\citet{bert-score} show that BERTScore correlation scores with humans drop as they select the last few layers of BERT for getting the embeddings. Therefore, the robustness of examined metrics may also depend on their corresponding selected layers for computing embeddings. 

In this section, we evaluate the impact of the selected hidden layer on the robustness of the metric. We evaluate three settings where we use: (a) the embeddings of the first layer for all models, (b) the embeddings of the best layer for each model (cf.\ Table \ref{tab:attack}), and (c) the mean of aggregated embeddings over all layers. We perform the robustness evaluations using the visual attack at $p=0.3$. Figure \ref{fig:why} shows the average results of this experiment\footnote{In Table \ref{tab:layer-no-attack} and \ref{tab:layer-vis-attack} in Appendix \ref{sect:impact}, we report scores for each language pair.}. We make the following observations.

First, using the embeddings of the first layer closes the gap between the correlations of different variations of the ByT5 model, i.e., small, base, and large, in the presence of more unknown tokens, i.e., $p=0.3$. 

Second, using the embeddings of the first layer improves the robustness of BERTScore using BERT embeddings, i.e., improving the correlation from 0.033 to 0.174 for BERT-base given $p=0.3$. However, the correlation of the resulting BERTScore is still considerably lower than using ByT5 embeddings at the presence of more unknown tokens. This indicates that \textbf{both} the choices of the hidden layer as well as the pretrained model play an important role in the robustness of the resulting embedding-based metric. 
{A reason why the first layer may be more effective in our setup is that, {in the presence of input noise or unknown tokens}, embeddings of higher layers may become less and less meaningful, as the noise may propagate and accumulate along layers. We provide an example from the similarity matrix of the resulting embeddings for different layers in Figure~\ref{fig:test} in the Appendix \ref{sect:effectiveness}.}

Overall, our results indicate that optimizing the layer on a standard data set such as WMT16 may be suboptimal in terms of the generalization of the resulting metrics. 
Concerning efficiency of the resulting metrics (a core aspect of modern NLP \citep{sustainlp-2020-sustainlp}), 
BERT-base has 110 million parameters, while ByT5-small has 300 million parameters. With the default BERTScore setting, passing the input through 9 layers results in a longer inference time. However, using the embeddings of the first layer results in a very fast inference for both models.

%% file: conclusion.tex
\section{Conclusion}
Embedding-based evaluation metrics will be used across different tasks and datasets that may contain data from very different domains. However, such metrics are only evaluated on standard datasets that contain similar domains as those used for pretraining embeddings. As a result, it is not clear how reliable the results of such evaluation metrics will be on new domains.
In this work, we investigate the robustness of embedding-based metrics in the presence of different ratios of unknown tokens. We show that (a) the results of the BERTScore using BERT-based embeddings is not robust, and its correlation with human evaluations drops significantly as the ratio of unknown tokens increases, and (b) using character-level embeddings from the first layer of ByT5 significantly improves the robustness of BERTScore and results in reliable results given different ratios of unknown tokens. 
We encourage the community to use this setting for their embedding-based evaluations, especially when applying the metrics to less standard domains.

In future work, we aim to address other aspects of robustness of evaluation metrics beyond an increased amount of unknown tokens as a result of spelling variation, such as how metrics cope with varying factuality \citep{Chen2022MENLIRE} or with fluency and grammatical acceptability issues \citep{rony-etal-2022-rome}. We also plan to investigate the impact of pixel-based representations \citep{Rust2022LanguageMW} (which are even more lower-level) for enhancing the robustness of evaluation metrics.  

\section*{Acknowledgments}
We thank the Ubiquitous Knowledge Processing (UKP) Lab and the system administrators of the lab for providing the computing resources to perform the experiments. We thank the anonymous reviewers for their comments and suggestions that improved the final version of the paper.

%% file: appendix.tex
\clearpage
\appendix
\section{Counting UNK token}
\label{sect:label_count}
\begin{figure*}[!htbp]
  \centering
  \begin{subfigure}{.49\linewidth}
    \centering
    \includegraphics[width=\linewidth]{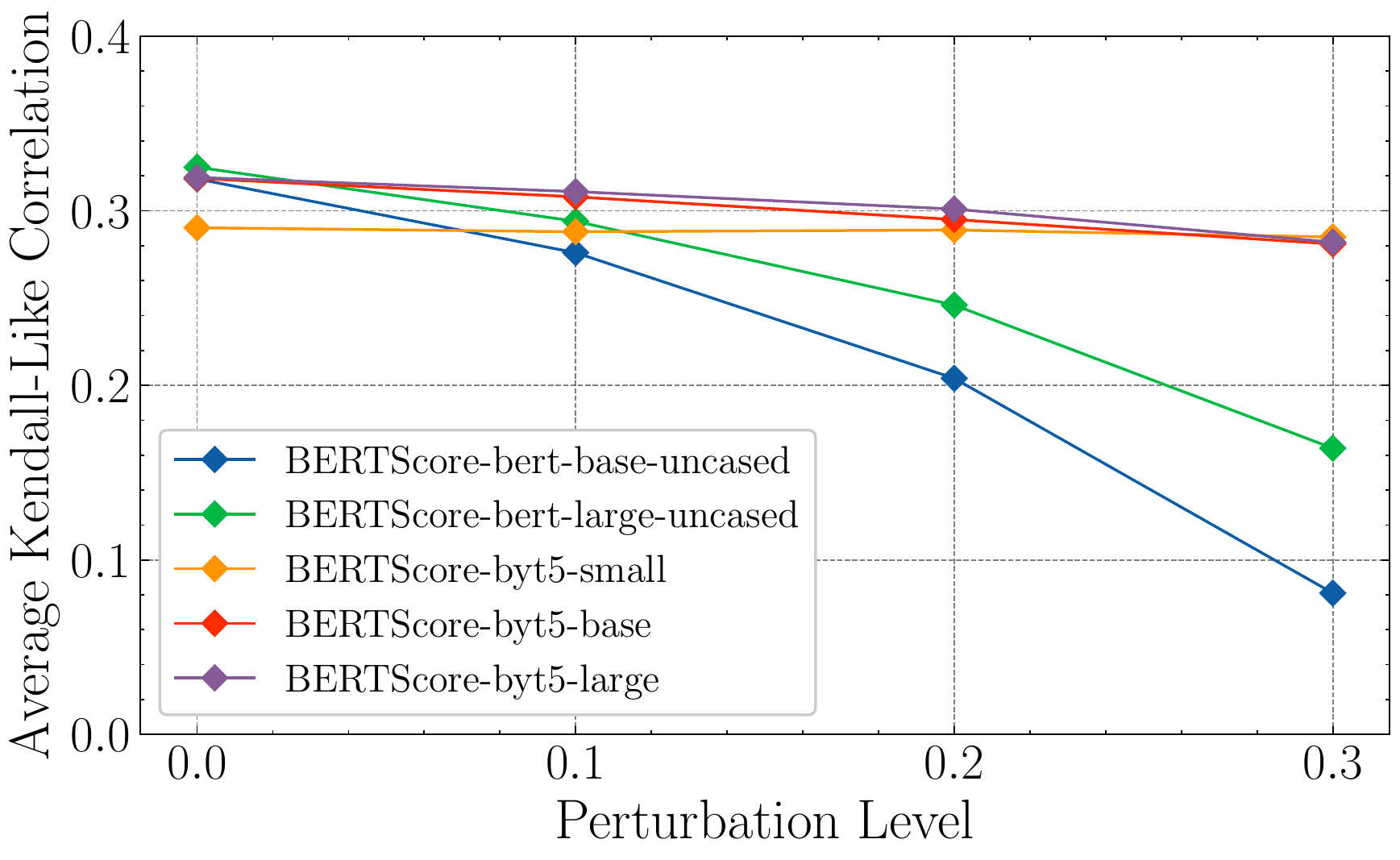}
    \caption{Intrude attack}
  \end{subfigure}
  \begin{subfigure}{.49\linewidth}
    \centering
    \includegraphics[width=\linewidth]{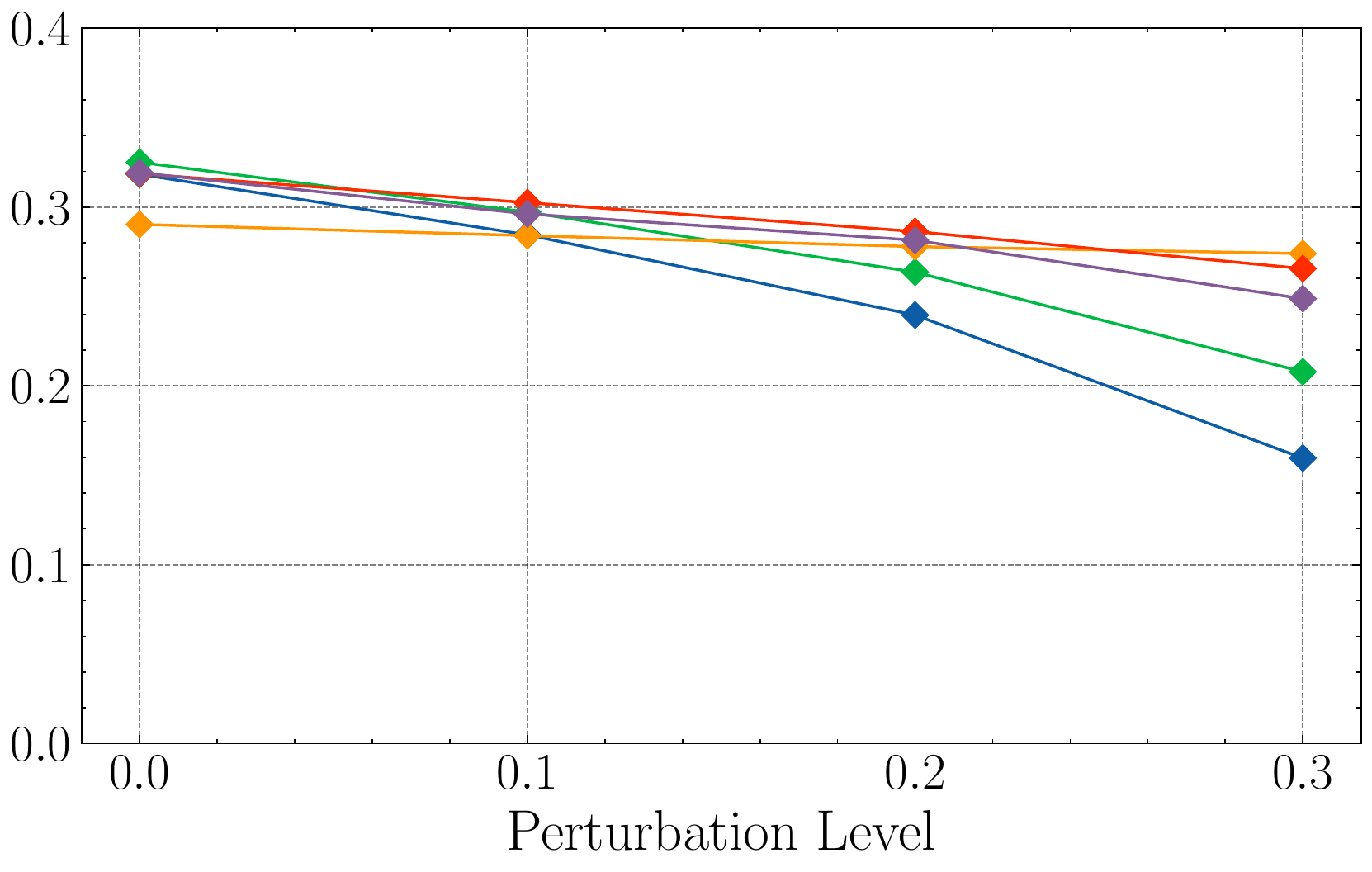}
    \caption{Disemvowel attack}
  \end{subfigure}\\
  \begin{subfigure}{.49\linewidth}
    \centering
    \includegraphics[width=\linewidth]{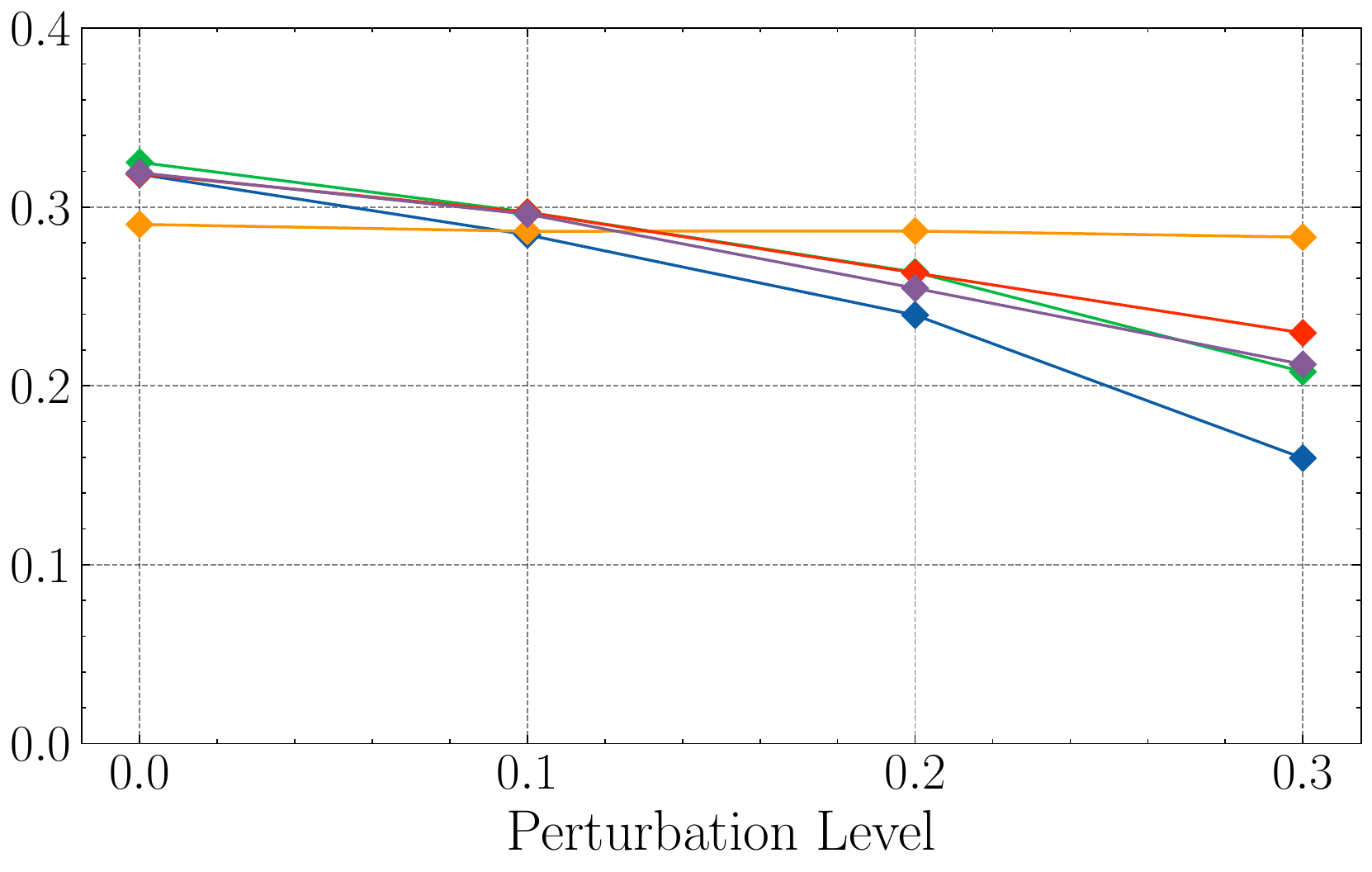}
    \caption{Phonetic attack}
  \end{subfigure}
  \begin{subfigure}{.49\linewidth}
    \centering
    \includegraphics[width=\linewidth]{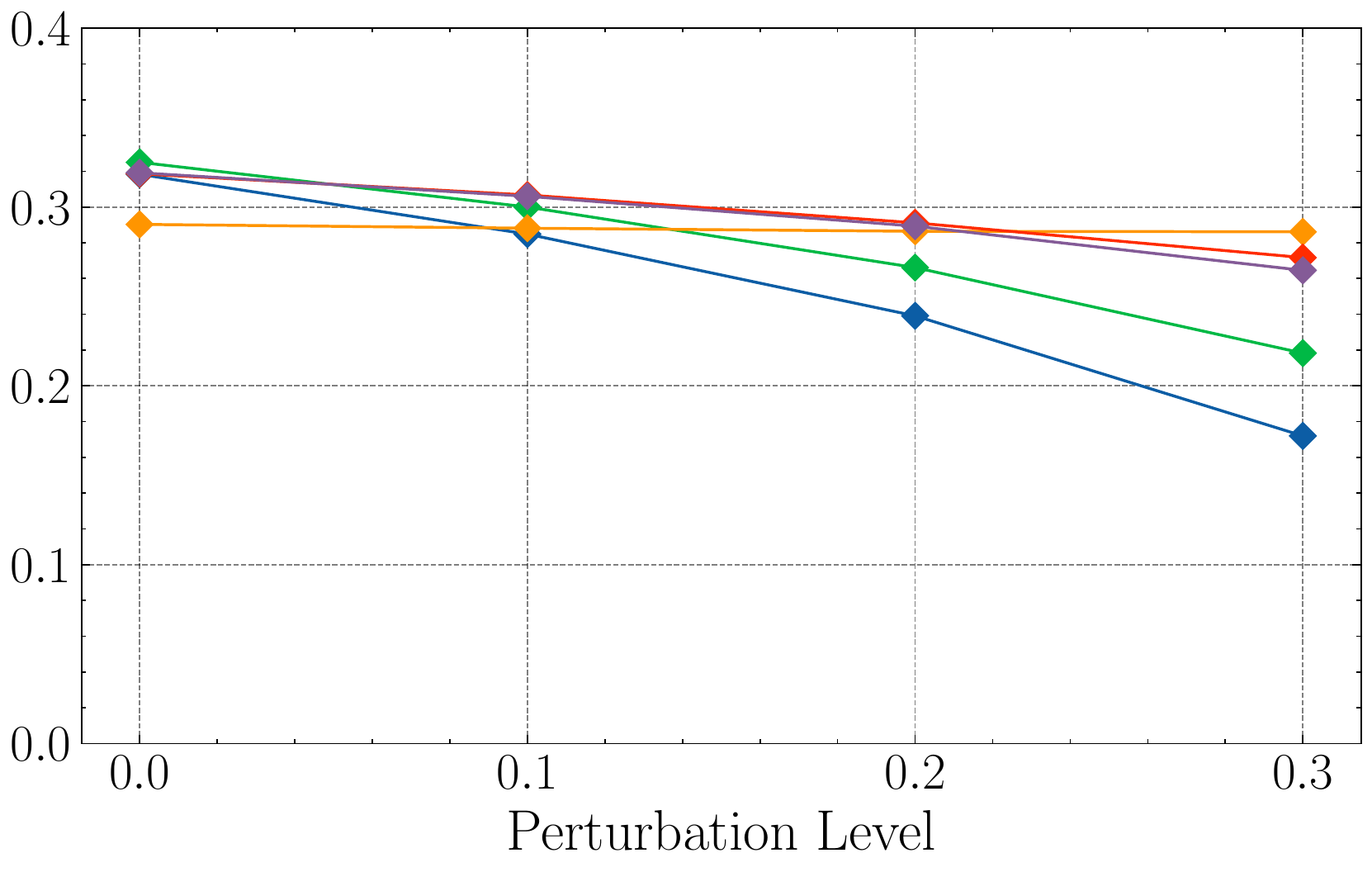}
    \caption{Keyboard-typo attack}
  \end{subfigure}
  \caption{Average Kendall correlation of seven to-English language pairs in WMT19 under attack with perturbation level from $p=0.0$ to $p=0.3$}
  \label{fig:wmt19-other}
\end{figure*}

\begin{algorithm}[t]
    \SetKwInOut{Input}{input}
    \SetKwInOut{Output}{output}
    \SetAlgoLined
    \caption{Count UNK token in a BERT tokenized sentence}
    \label{count_unk}
    \SetKwProg{Def}{def}{:}{}
    \Def{count\_UNK}{
        \KwData{\emph{sentence}: a tokenized sentence as a list of string}
        \Output{\emph{count}: number of UNK token of input tokenized sentence}
        \BlankLine
        \textit{count} $\longleftarrow 0$\\
        \textit{buffer} $\longleftarrow$ \textbf{empty} list\\
        \For{\textit{token} \textbf{in} \textit{sentence}}{
            \uIf{\text{[UNK]} \textbf{in} \textit{token}}{
                \textit{count} $\longleftarrow$ \textit{count} $+ 1$
            }
            \uElseIf{\#\# \textbf{in} token}{
                \textbf{Add} \textit{token} \textbf{to} \textit{buffer}
            }
            \Else{
                \If{len(buffer) != 0}{
                    \textit{count} $\longleftarrow$ \textit{count} $+ 1$\\
                    \textbf{Empty} \textit{buffer}
                }
            }
        }
        \If{len(sentence) $\geq$ 2}{
            \If{\#\# \textbf{in} last token \textbf{of} sentence}{
                \textit{count} $\longleftarrow$ \textit{count} $+ 1$
            }
        }
        \Return{\textit{count}}
    }
\end{algorithm}

Algorithm \ref{count_unk} shows how we count UNK tokens that the BERT tokenizer creates from a sentence. In BERT, \texttt{[UNK]} represents the UNK tokens that are not in their given vocabulary. Besides \texttt{[UNK]}, BERT use WordPiece tokenizer concept, which breaks the unknown word into sub-words using a greedy longest-match-first algorithm, such as splits ``\emph{bassing}'' into `\emph{bass}' and `\emph{\#\#ing}' where `\emph{\#\#}' denotes the join of sub-words. Thus, the UNK word becomes two known words. `\emph{\#\#}' is the indication for the starting of a UNK word if the previous token does not contain `\emph{\#\#}'. In case the next token still contains `\emph{\#\#}', it indicates that the token still belongs to a word and does not count as a UNK token, e.g., ``\emph{verständlich}'' to \emph{`vers', `\#\#tä', `\#\#nd', `\#\#lich'} and count it as one UNK token. It lasted until we finally found non contain `\emph{\#\#}' token. With a word-piece tokenizer, the beginning token of a tokenized sentence is either \texttt{[UNK]} or known word, and we also consider the case where the last token contains ``\emph{\#\#}''.

\begin{table}[t]
  \centering
  \begin{tabular}{@{}l|r@{}}
  \toprule
  Language Pair               & No. Segment \\ \midrule
  bn-hi (Bengali $\to$ Hindi) & 4,461       \\
  hi-bn (Hindi $\to$ Bengali) & 4,512       \\
  xh-zu (Xhosa $\to$ Zulu)    & 2,952       \\
  zu-xh (Zulu $\to$ Xhosa)    & 2,502       \\ \bottomrule
  \end{tabular}
  \caption{Amount of segments in WMT21 for Hindi $\longleftrightarrow$ Bengali and Zulu $\longleftrightarrow$ Xhosa.}
  \label{tab:flores}
\end{table}

\section{WMT19}
\label{sect:wmt19}
The results of other attacks are illustrated in Figure~\ref{fig:wmt19-other}.

\begin{table*}[t]
    \centering
    {\small
    \begin{tabular}{@{}llcccccccc@{}}
    \toprule
    Setting                  & Metric                      & de-en          & fi-en          & gu-en          & kk-en          & lt-en          & ru-en          & zh-en          & Average        \\ \midrule
    \multirow{4}{*}{Default} & BERTScore-bert-base-uncased & 0.18           & 0.339          & 0.288          & 0.438          & 0.364          & 0.209          & 0.41           & 0.318          \\
                             & BERTScore-byt5-small        & 0.172          & 0.286          & 0.278          & 0.422          & 0.307          & 0.194          & 0.373          & 0.290          \\
                             & BERTScore-byt5-base         & \textbf{0.197} & 0.326          & 0.297          & 0.419          & 0.358          & \textbf{0.215} & \textbf{0.418} & \textbf{0.319} \\
                             & BERTScore-byt5-large        & 0.193          & 0.333          & 0.304          & 0.427          & 0.354          & 0.208          & 0.415          & \textbf{0.319} \\ \midrule
    \multirow{4}{*}{First}   & BERTScore-bert-base-uncased & 0.147          & 0.295          & 0.263          & 0.421          & 0.318          & 0.183          & 0.361          & 0.284          \\
                             & BERTScore-byt5-small        & 0.171          & 0.285          & 0.279          & 0.422          & 0.307          & 0.194          & 0.370           & 0.290          \\
                             & BERTScore-byt5-base         & 0.164          & 0.276          & 0.280           & 0.414          & 0.307          & 0.191          & 0.362          & 0.285          \\
                             & BERTScore-byt5-large        & 0.161          & 0.277          & 0.280           & 0.416          & 0.308          & 0.189          & 0.361          & 0.285          \\ \midrule
                             & BERTScore-bert-base-uncased & 0.17           & \textbf{0.326} & 0.289          & \textbf{0.437} & \textbf{0.351} & 0.206          & 0.397          & 0.311          \\
    Mean of                  & BERTScore-byt5-small        & 0.170           & 0.292          & 0.284          & 0.420           & 0.313          & 0.202          & 0.372          & 0.293          \\
    aggregation              & BERTScore-byt5-base         & 0.188          & 0.324          & \textbf{0.305} & 0.427          & 0.347          & 0.207          & 0.408          & 0.315          \\
                             & BERTScore-byt5-large        & 0.185          & 0.322          & 0.311          & 0.431          & 0.343          & 0.208          & 0.411          & 0.316          \\ \bottomrule
    \end{tabular}}
    \caption{Segment-level correlation metric results Kendall for seven to-non-English language pairs in WMT19 with respect to fist layer, default layer and mean of aggregated embeddings setting without any attack  i.e. $p=0$.}
    \label{tab:layer-no-attack}
\end{table*}

\begin{table*}[t]
    \centering
    {\small
    \begin{tabular}{@{}llcccccccc@{}}
    \toprule
    Setting                  & Metric                      & \multicolumn{1}{c}{de-en} & \multicolumn{1}{c}{fi-en} & \multicolumn{1}{c}{gu-en} & \multicolumn{1}{c}{kk-en} & \multicolumn{1}{c}{lt-en} & \multicolumn{1}{c}{ru-en} & \multicolumn{1}{c}{zh-en} & \multicolumn{1}{c}{Average} \\ \midrule
    \multirow{4}{*}{Default} & BERTScore-bert-base-uncased & -0.003                    & -0.014                    & -0.027                    & 0.149                     & -0.022                    & 0.024                     & 0.126                     & 0.033                       \\
                             & BERTScore-byt5-small        & \textbf{0.155}            & \textbf{0.266}            & 0.239                     & 0.392                     & \textbf{0.264}            & \textbf{0.175}            & \textbf{0.360}             & \textbf{0.264}              \\
                             & BERTScore-byt5-base         & 0.014                     & -0.009                    & 0.026                     & 0.147                     & 0.052                     & 0.042                     & 0.155                     & 0.061                       \\
                             & BERTScore-byt5-large        & 0.011                     & -0.055                    & -0.018                    & 0.141                     & -0.015                    & 0.032                     & 0.155                     & 0.036                       \\ \midrule
    \multirow{4}{*}{First}   & BERTScore-bert-base-uncased & 0.074                     & 0.215                     & 0.082                     & 0.215                     & 0.234                     & 0.120                      & 0.278                     & 0.174                       \\
                             & BERTScore-byt5-small        & \textbf{0.155}            & \textbf{0.266}            & 0.239                     & 0.392                     & \textbf{0.264}            & \textbf{0.175}            & \textbf{0.360}             & \textbf{0.264}              \\
                             & BERTScore-byt5-base         & 0.147                     & 0.256                     & \textbf{0.262}            & \textbf{0.403}            & 0.264                     & 0.166                     & 0.348                     & \textbf{0.264}              \\
                             & BERTScore-byt5-large        & 0.138                     & 0.258                     & 0.259                     & 0.394                     & 0.262                     & 0.170                      & 0.352                     & 0.262                       \\ \midrule
                             & BERTScore-bert-base-uncased & 0.053                     & 0.144                     & 0.052                     & 0.214                     & 0.149                     & 0.082                     & 0.240                      & 0.133                       \\
    Mean of                  & BERTScore-byt5-small        & 0.070                      & 0.089                     & 0.094                     & 0.244                     & 0.109                     & 0.107                     & 0.273                     & 0.141                       \\
    aggregation              & BERTScore-byt5-base         & 0.025                     & -0.029                    & 0.022                     & 0.263                     & -0.019                    & 0.056                     & 0.123                     & 0.063                       \\
                             & BERTScore-byt5-large        & 0.054                     & 0.005                     & 0.020                      & 0.255                     & 0.013                     & 0.095                     & 0.156                     & 0.085                       \\ \bottomrule
    \end{tabular}}
    \caption{Segment-level correlation metric results Kendall for seven to-non-English language pairs in WMT19 with respect to fist layer, default layer and mean of aggregated embeddings setting under visual attack at 0.3 perturbation level.}
    \label{tab:layer-vis-attack}
\end{table*}

 \section{FLORES}
 \label{sect:flores}
 Table \ref{tab:flores} shows the number of provided human annotations in FLORES.

 \section{Impact of layer choice in BERTScore} \label{sect:impact}
Table \ref{tab:layer-no-attack} and \ref{tab:layer-vis-attack} show the particular results of each language pair with different settings in BERTScore without attack and with visual attack at $p=0.3$ respectively. 

\section{Effectiveness of the first layer} \label{sect:effectiveness}
\begin{figure*}
    \centering
    \begin{subfigure}{.5\linewidth}
    \centering
        \includegraphics[width=\linewidth]{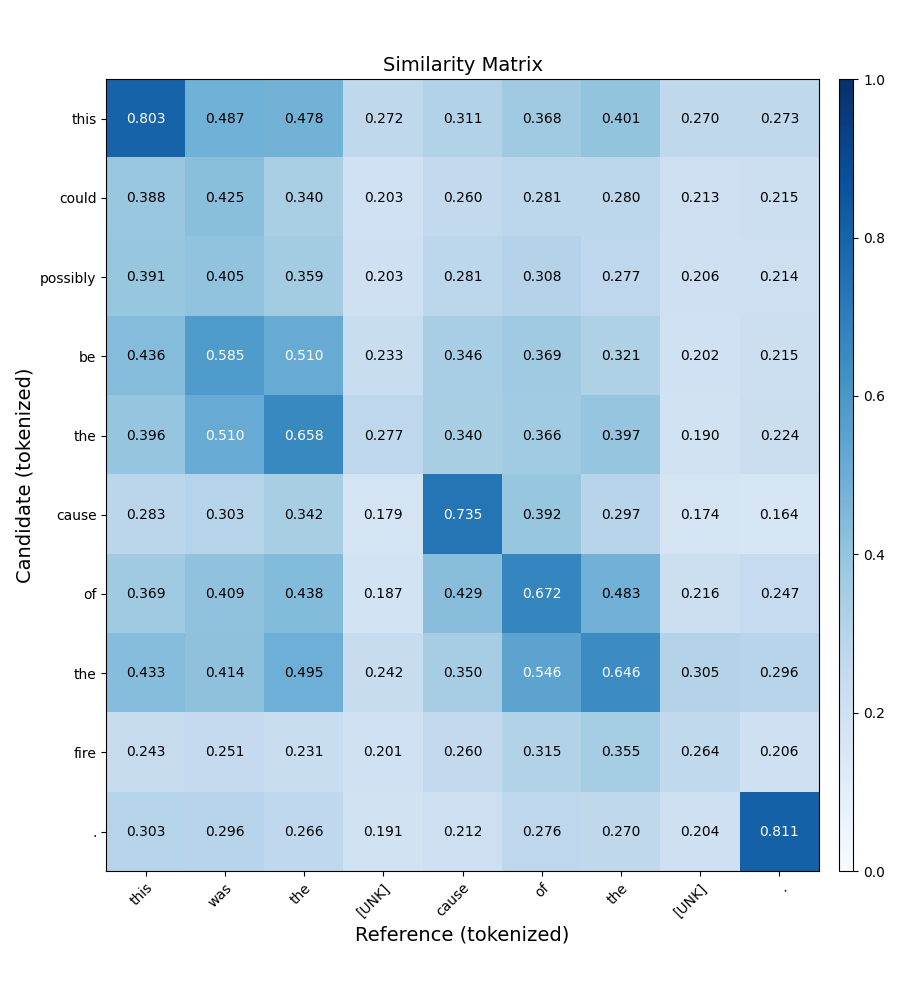}
        \caption{9th layer, \emph{attacked reference}: \\``\scalerel*{\includegraphics{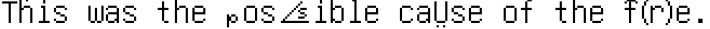}}{15pt} ''}
        \label{fig:sub1}
    \end{subfigure}%
    \begin{subfigure}{.5\linewidth}
        \centering
        \includegraphics[width=\linewidth]{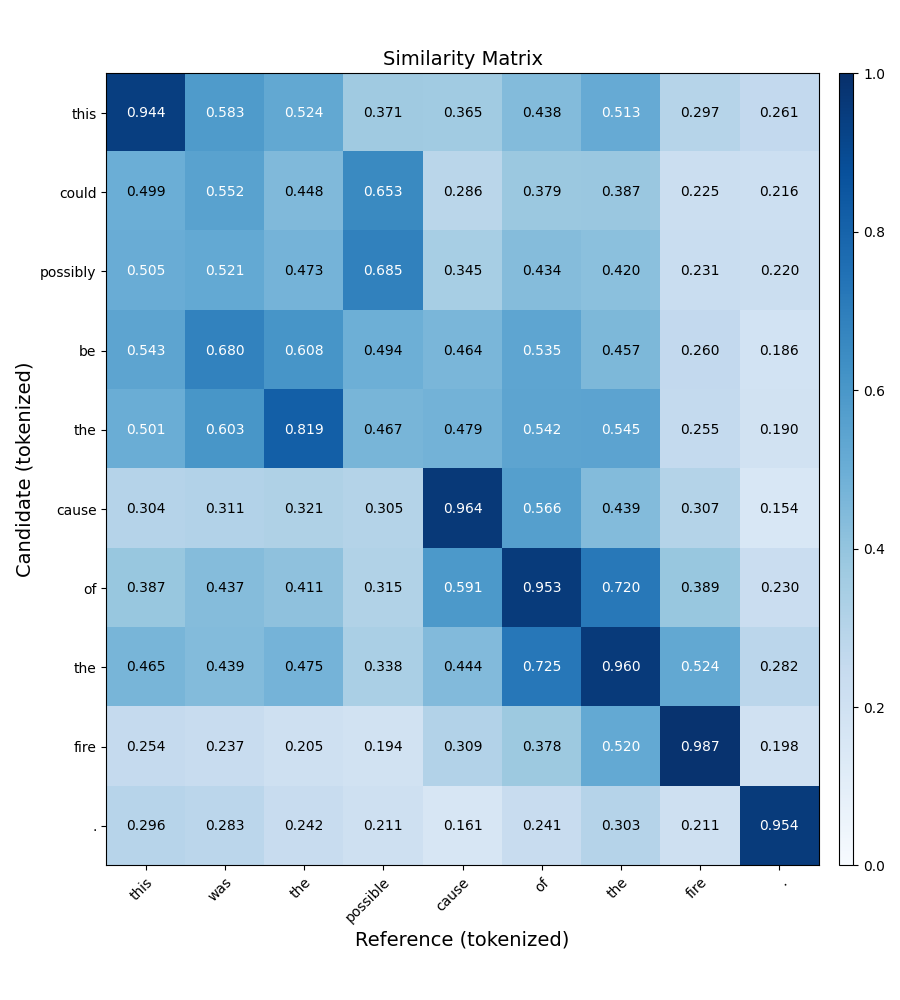}
        \caption{9th layer, \emph{normal reference}: \\``This could possibly be the cause of the fire.''}
        \label{fig:sub2}
    \end{subfigure}\\[1ex]
    \begin{subfigure}{.5\linewidth}
        \centering
        \includegraphics[width=\linewidth]{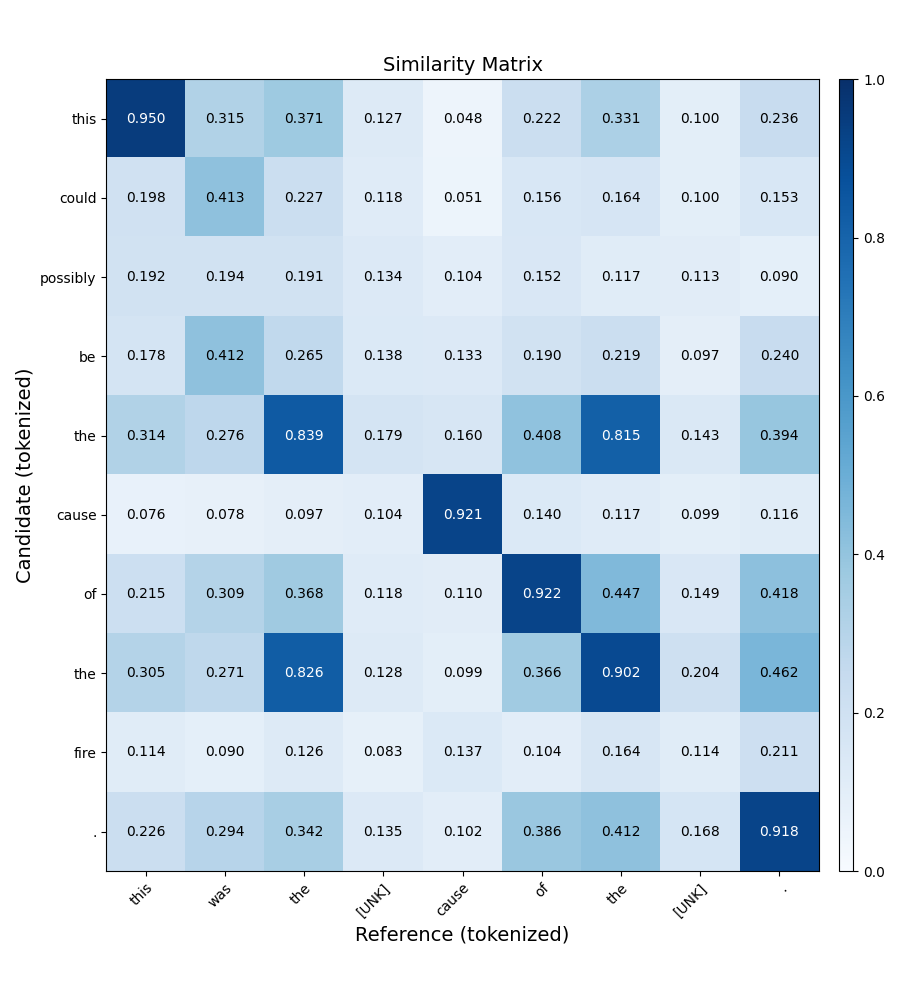}
        \caption{1st layer, \emph{attacked reference}: \\``\scalerel*{\includegraphics{appendix-attacked-reference.pdf}}{15pt}''}
        \label{fig:sub3}
    \end{subfigure}%
    \begin{subfigure}{.5\linewidth}
        \centering
        \includegraphics[width=\linewidth]{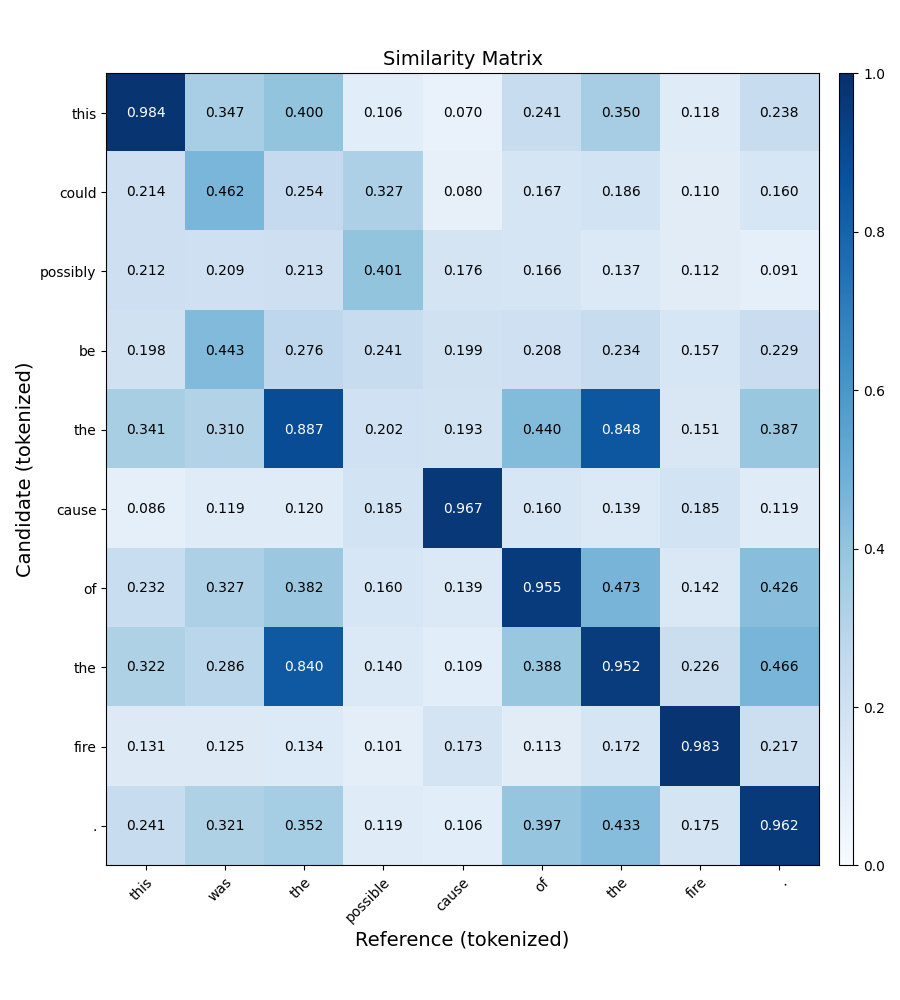}
        \caption{1th layer, \emph{normal reference}: \\`` This could possibly be the cause of the fire.''}
        \label{fig:sub4}
    \end{subfigure}
    \caption{Similarity Matrix using BERTScore with bert-base-uncased for \emph{candidate}: `` This could possibly be the cause of the fire.'' in different setups.}
    \label{fig:test}
\end{figure*}

In Figure \ref{fig:test}, we show four different settings and their cosine similarity matrix computed by BERTScore using bert-base-uncased. In both \emph{normal reference} with 1st or 9th setups, matched tokens get higher similarity score. 9th layer setting gathers information for relevant tokens, which results in higher similarity score across the matrix. As in the case with \emph{attacked reference}, 1st layer setting penalizes the unmatched tokens and the magnitude of matched tokens are as high as using \emph{normal reference} with 1st layer setup. However, by using 9th layer for \emph{attacked reference}, we can observe the hue color of matched tokens with low score. Thus, we conclude the accumulated noise to higher layer cause the problem with effectiveness in our previous setup with WMT19 dataset.
